# Collaborative Training of Medical Artificial Intelligence Models with non-uniform Labels

Soroosh Tayebi Arasteh (1), Peter Isfort (1), Marwin Saehn (1), Gustav Mueller-Franzes (1), Firas Khader (1), Jakob Nikolas Kather (2,3,4,5), Christiane Kuhl (1), Sven Nebelung* (1), Daniel Truhn* (1)

(1) Department of Diagnostic and Interventional Radiology, University Hospital RWTH Aachen, Aachen, Germany.
(2) Department of Medicine III, University Hospital RWTH Aachen, Aachen, Germany.
(3) Else Kroener Fresenius Center for Digital Health, Medical Faculty Carl Gustav Carus, Technical University Dresden, Dresden, Germany.
(4) Division of Pathology and Data Analytics, Leeds Institute of Medical Research at St James's, University of Leeds, Leeds, UK.
(5) Medical Oncology, National Center for Tumor Diseases (NCT), University Hospital Heidelberg, Heidelberg, Germany.

**Abstract:**
Due to the rapid advancements in recent years, medical image analysis is largely dominated by deep learning (DL). However, building powerful and robust DL models requires training with large multi-party datasets. While multiple stakeholders have provided publicly available datasets, the ways in which these data are labeled vary widely. For Instance, an institution might provide a dataset of chest radiographs containing labels denoting the presence of pneumonia, while another institution might have a focus on determining the presence of metastases in the lung. Training a single AI model utilizing all these data is not feasible with conventional federated learning (FL). This prompts us to propose an extension to the widespread FL process, namely flexible federated learning (FFL) for collaborative training on such data. Using 695,000 chest radiographs from five institutions from across the globe - each with differing labels - we demonstrate that having heterogeneously labeled datasets, FFL-based training leads to significant performance increase compared to conventional FL training, where only the uniformly annotated images are utilized. We believe that our proposed algorithm could accelerate the process of bringing collaborative training methods from research and simulation phase to the real-world applications in healthcare.

**Keywords:**
Artificial Intelligence, Chest X-ray, Federated Learning, Medical Image Analysis, Multilabel Disease Classification

# Introduction

Artificial Intelligence (AI) is widely expected to reshape medicine in the next decade[1]. The development of robust and clinically useable AI models hinges however on the availability of large and multi-institutional datasets as illustrated by recent publications that have advanced the field in many different areas covering diagnosis and prognosis of diseases in radiological[2,3] and histopathological[4–6] use-cases. One solution to use multi-institutional datasets is conventional federated learning (FL)[7–9] in which the AI model is sent to multiple collaborating centers for training. However, this paradigm requires that the model sees data that is labeled in exactly the same way at each center, i.e., if one center has labeled the presence of pneumonia in its dataset, all the other participating centers also need to label their data with the presence of pneumonia[2,10–13]. While these requirements can be met if the study is carefully planned before the start of data acquisition, in more realistic scenarios, centers often already possess large data that has been individually labeled. In medicine in particular, labels might differ quite dramatically, since the labeling process is complex and since there is no standardized way of labeling the presence of a disease[14–16]. Labels might often be created by two different centers and might be closely related yet appear completely separate to the algorithm that is to be trained. For example, center A might have annotated a dataset of thoracic radiographs with binary labels about the presence of cardiomegaly, while center B might have decided to label another dataset of thoracic radiographs with binary labels about the presence of lung congestion. Both labeling schemes are related and there is mutual information in the labels, since patients with an enlarged heart are more prone to lung congestion, however, conventional FL does not allow to jointly train a model with these data[17].

     In this study, we propose flexible federated learning (FFL) as a solution to this impediment on collaboration. In our architecture we divide the classification network into a classification head and a feature extraction backbone. The backbone is shared between all sites and weights are jointly trained in a FL scheme. The classification head on the other hand can be tailored to the local data with an individual loss function, see **Fig. 1**. Our goal was to collaboratively and securely train a common backbone network using all data from separate data owners utilizing all available labels. Our hypothesis was that this backbone network learns to extract features that are relevant for the classification of related, but different tasks and that using such a common - and jointly trained - backbone improves classification accuracy for each participating center. We tested this hypothesis on five multicentric datasets comprising a total of over 695,000 thoracic radiographs. The labels assigned to the radiographs from each of the five centers differed, but were related and carried similar information content, thus providing the ideal testing ground for our paradigm.

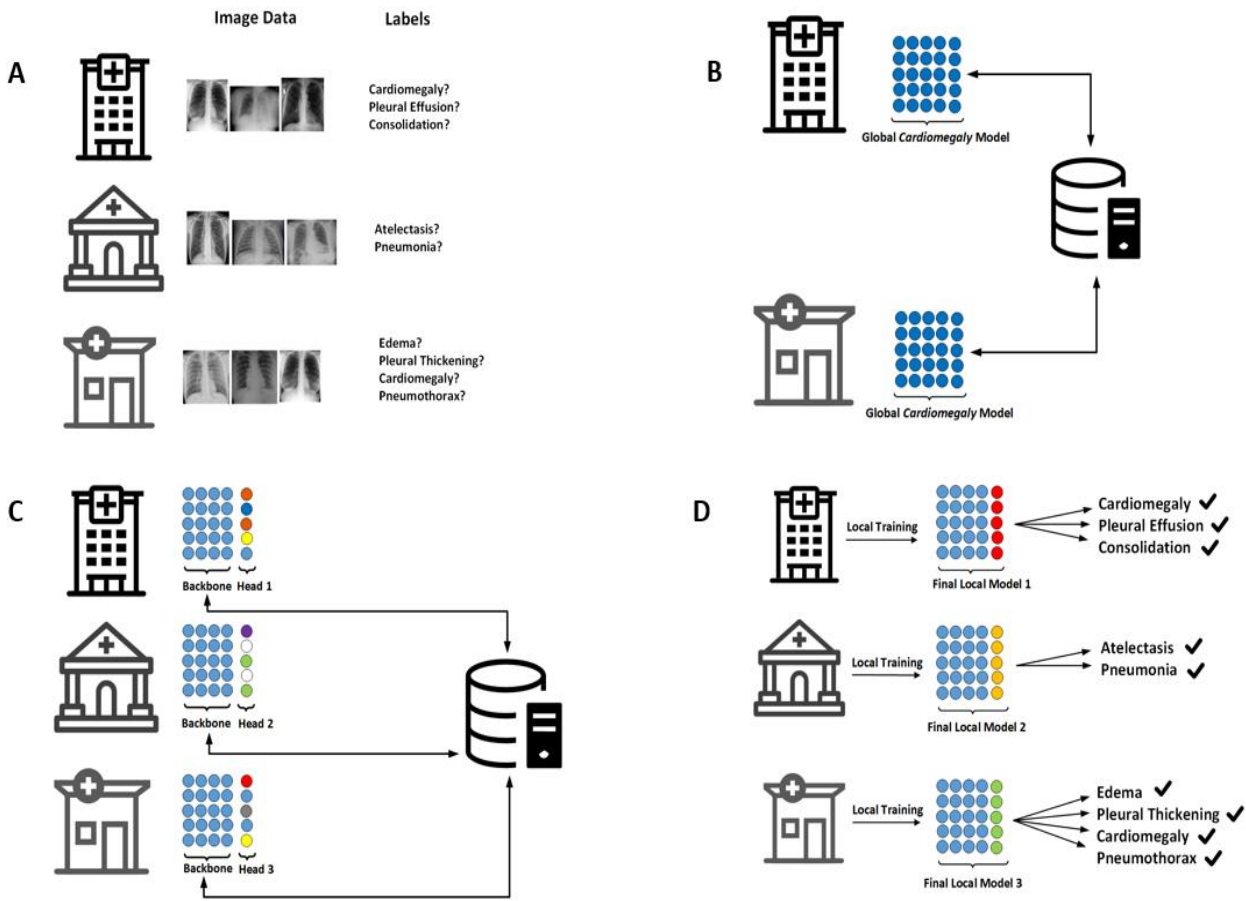

**Fig. 1**: Overview of the flexible federated learning (FFL) process. **(A)** Three separate data centers intend to train AI models for the prediction of different diseases. **(B)** Conventional federated learning: only center 1 and center 3 who have overlapping objectives can collaborate on training a neural network for the detection of cardiomegaly only. **(C)** FFL: all centers collaborate to train a common backbone network and individual classification heads using all their data. **(D)** For classification, each center employs the common backbone and the local classification head.

## Results

### FFL Trains Medical Classification Models with superior Performance on Non-Overlapping Labels

We first test our hypothesis that FFL performs superior to conventional FL in a prototypical setting with high-quality data. We utilized two datasets that were both manually labeled by expert radiologists: VinDr-CXR[18,19], a public dataset of thoracic radiographs and UKA-CXR, a private dataset of intensive care thoracic radiographs[20]. Labels for both datasets were different, such that training in a conventional FL setting was not possible. In particular, UKA-CXR has labels for a dedicated set of pathologies for each patient side (e.g., left lung and right lung), while VinDr-CXR utilizes a different set of pathologies and global labels (indicating the presence of a disease in the left *or* the right lung), see **Table 1** and **Fig. 2A**. We chose two distinct label categories in each dataset that have overlapping information content: *cardiomegaly* and *pleural effusion* for VinDr-CXR and *right pleural effusion* and *left pneumonic* *infiltrates* for the UKA-CXR dataset. Subsequently we trained a ResNet[21] within our FFL scheme on the full UKA-CXR dataset (n=122,294 training images) and on varying amounts of data from VinDr-CXR (n=2,000, 5,000, and 15,000). When tested on a held-out benchmark test set of VinDr-CXR, the average area under the receiver-operator-curve (AUROC) was significantly higher when applying FFL as compared to local training (0.90 ± 0.02 vs. 0.86 ± 0.04; p=0.001). We observed a similar trend when increasing the training set to n=5,000 (0.92 ± 0.02 vs. 0.90 ± 0.01; p=0.003) and n=15,000, i.e., the full dataset (0.95 ± 0.01 vs. 0.94 ± 0.02; p=0.035). Thus, in all of these experiments, FFL improved performance as compared to local training.

**Table 1**: Results of the comparison between local and FFL-based training of VinDr-CXR dataset with non-overlapping labels for different training set sizes, tested on the VinDr-CXR benchmark. Average area under the receiver-operator-curve (AUROC) over *cardiomegaly* and *pleural effusion*. The FFL was performed in combination with UKA-CXR dataset of n=122,294 images with two different labels including *pleural effusion right* and *pneumonic infiltrates right*.

|  | Local VinDr 2K | FFL VinDr 2K | Local VinDr 5K | FFL VinDr 5K | Local VinDr 15K | FFL VinDr 15K |
| --- | --- | --- | --- | --- | --- | --- |
| AUROC | 0.86 ± 0.04 | 0.90 ± 0.02 | 0.90 ± 0.01 | 0.92 ± 0.02 | 0.94 ± 0.02 | 0.95 ± 0.01 |
| P-value | 0.001 | | 0.003 | | 0.035 | |

## FFL Trains Medical Classification Models with superior Performance on Partly Overlapping Labels

Next, we extended the available classification labels to comprise seven categories in each dataset. Part of these labels overlap, e.g., *cardiomegaly*, while others again denote distinct categories. This reflects a more realistic scenario in which both sites have independently labeled their data on common pathologies but differ in the details of their labeling approach. In particular, for the VinDr-CXR dataset we employ the labels *no finding*, *aortic enlargement*, *pleural thickening*, *cardiomegaly*, *pleural effusion*, *pneumothorax*, and *atelectasis* and for the UKA-CXR dataset we employ *cardiomegaly*, *pleural effusion right*, *pleural effusion left*, *pneumonic infiltrates right*, *pneumonic infiltrates left*, *atelectasis right*, and *atelectasis left*.

By analogy with the first experiment, we compared local training to FFL-based training for subsets of n=2,000, n=5,000, and n=15,000 labeled radiographs. Again, when tested on the held-out benchmark test set of VinDr-CXR, the average AUROC was higher when applying FFL as compared to local training (0.78 ± 0.06 vs. 0.77 ± 0.08; p=0.340). Similar results were observed when increasing the training set to n=5,000 (0.82 ± 0.05 vs. 0.79 ± 0.07; p=0.010) and n=15,000, i.e., the full dataset (0.84 ± 0.05 vs. 0.83 ± 0.09; p=0.180), see **Table 2** and **Fig. 2B**. Thus, FFL improves performance of classification models on partly overlapping data as compared to local training.

**Table 2**: Results of the comparison between local and FFL-based training of VinDr-CXR dataset with overlapping labels for different training set sizes, tested on the VinDr-CXR benchmark. Average AUROC values over *no finding*, *aortic enlargement*, *pleural thickening*, *cardiomegaly*, *pleural effusion*, *pneumothorax*, and *atelectasis*. The FFL was performed in combination with UKA-CXR dataset of n=122,294 images with 7 other labels including *cardiomegaly*, pleural *effusion right*, *pleural effusion left*, *pneumonic infiltrates right*, *pneumonic infiltrates left*, *atelectasis right*, and *atelectasis left*.

|  | Local VinDr 2K | FFL VinDr 2K | Local VinDr 5K | FFL VinDr 5K | Local VinDr 15K | FFL VinDr 15K |
|---|---|---|---|---|---|---|
| AUROC | 0.77 ± 0.08 | 0.78 ± 0.06 | 0.79 ± 0.07 | 0.82 ± 0.05 | 0.83 ± 0.09 | 0.84 ± 0.05 |
| P-value | 0.340 | | 0.010 | | 0.180 | |

## FFL is Scalable

To examine if FFL keeps its advantageous properties when trained on truly large and diverse multi-centric datasets, we perform the following experiment: we employ five independent cohorts of thoracic radiographs who each are trained on five labels: 1) the VinDr-CXR dataset (n=15,000) with labels including *no finding, aortic enlargement, pleural thickening, cardiomegaly,* and *pleural effusion*; 2) the ChestX-ray14[22] dataset (n=86,524) with labels including *cardiomegaly, effusion, pneumonia, consolidation,* and *no finding;* 3) the CheXpert[23] dataset (n=128,356) with labels including *cardiomegaly, lung opacity, lung lesion, pneumonia,* and *edema;* 4) the MIMIC-CXR[24,25] dataset (n=210,652) with labels including *enlarged cardiomediastinum, consolidation, pleural effusion, pneumothorax,* and *atelectasis;* and 5) the UKA-CXR dataset (n=122,294) with labels including *pleural effusion left, pleural effusion right, cardiomegaly, pneumonic infiltrates left,* and *pneumonic infiltrates right*. It should be noted that only the UKA-CXR and the VinDr-CXR dataset have labels that were manually set by medical experts, while the remaining three datasets have labels extracted from natural language processing of radiological reports. For each of the five cohorts, we performed local training and compared it to training within our FFL framework for hold-out test set of each cohort. In all cohorts, FFL-based training outperformed local training in terms of the average AUROC (VinDr-CXR: 0.885 ± 0.049 vs. 0.867 ± 0.045, p=0.001; ChestX-ray14: 0.744 ± 0.080 vs. 0.744 ± 0.076, p=0.363; CheXpert: 0.797 ± 0.061 vs. 0.796 ± 0.064, p=0.243; MIMIC-CXR: 0.786 ± 0.066 vs. 0.772 ± 0.072, p=0.004; UKA-CXR 0.918 ± 0.031 vs. 0.916 ± 0.031; p=0.001, respectively), see **Table 3** and **Fig. 2C**. Thus, even though we observe a saturation effect if the local data comprises thousands of thoracic radiographs, FFL improves performance as compared to local training and can still be used if the data is labeled with vastly different labeling regimes.

**Table 3**: Results of the comparison between local and FFL-based training for 5 different datasets. Average AUROC values over all included labels for each dataset, tested on the test benchmark of the corresponding dataset. The FFL process for each dataset was performed in combination with the other 4 datasets including 5 different labels for each dataset.

| Dataset name | Training set size | Included labels | Training setup | AUROC | P-value |
|---|---|---|---|---|---|
| VinDr-CXR | n=15,000 | *no finding, aortic enlargement, pleural thickening, cardiomegaly, pleural effusion* | Local | 0.867 ± 0.045 | 0.001 |
| | | | FFL | 0.885 ± 0.049 | |
| ChestX-ray14 | n=83,525 | *cardiomegaly, lung opacity, lung lesion, pneumonia, edema* | Local | 0.744 ± 0.076 | 0.363 |
| | | | FFL | 0.744 ± 0.080 | |
| CheXpert | n=126,141 | *cardiomegaly, lung opacity, lung lesion, pneumonia, edema* | Local | 0.796 ± 0.064 | 0.243 |
| | | | FFL | 0.797 ± 0.061 | |
| MIMIC-CXR-JPG-v2.0 | n=237,972 | *enlarged cardiomediastinum, consolidation, pleural effusion, pneumothorax, atelectasis* | Local | 0.772 ± 0.072 | 0.004 |
| | | | FFL | 0.786 ± 0.066 | |
| UKA-CXR | n=122,297 | *pleural effusion left, pleural effusion right, cardiomegaly, pneumonic infiltrates left, pneumonic infiltrates right* | Local | 0.916 ± 0.031 | 0.001 |
| | | | FFL | 0.918 ± 0.031 | |

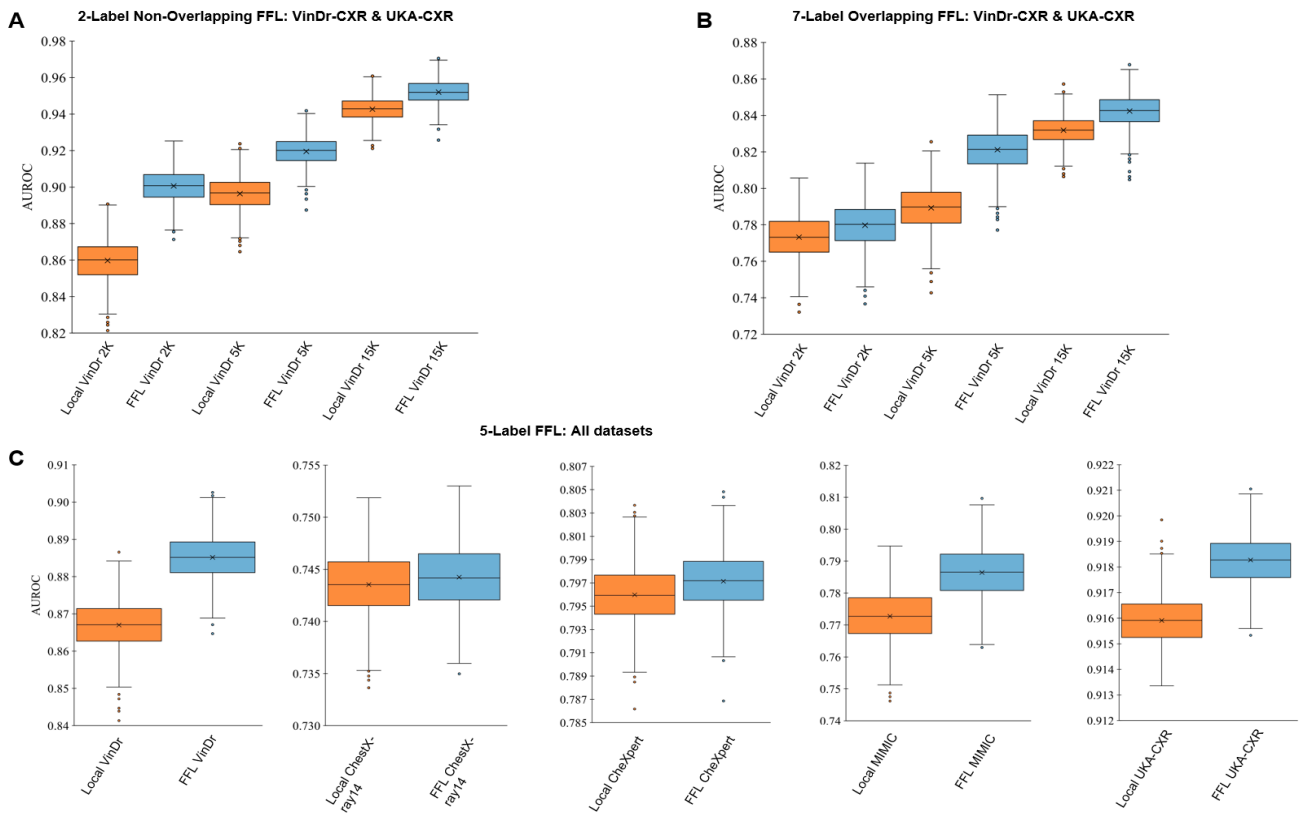

**Fig. 2**: Comparison between flexible federated learning (FFL)-based training and local training of classification models. **(A)** FFL-based training on UKA-CXR data (n=122,294, labels: *pleural effusion right* and *pneumonic infiltrates right*) and on VinDr-CXR data (on 2K, 5K and 15K images, labels: *cardiomegaly* and *pleural effusion*) if there is no overlap between labels. Performance tested on an independent VinDr-CXR test set. **(B)** Same setup as in (A), but training is performed with partially overlapping labels on UKA-CXR (n=122,294, labels: *cardiomegaly*, *pleural effusion right*, *pleural effusion left*, *pneumonic infiltrates right*, *pneumonic infiltrates left*, *atelectasis right*, and *atelectasis left*) and on VinDr-CXR (on 2K, 5K and 15K images, labels:  over *no finding*, *aortic enlargement*, *pleural thickening*, *cardiomegaly*, *pleural effusion*, *pneumothorax*, and *atelectasis*). **(C)** FFL-based training on five different datasets (VinDr-CXR, n=15,000; ChestX-ray14, n=86,524; CheXpert, n=128,356; MIMIC-CXR, n=210,652; and UKA-CXR, n=122,294). Testing is performed on the respective held-out test data.

## Discussion

AI models are becoming increasingly important in modern medicine and are currently reaching a stage in which they can improve patient care and render medical processes more efficient[26–35]. However, the biggest limitation in the development of such data-driven AI models, is their need to access large amounts of annotated data for training. For this, stakeholders need to be able to collaborate on a large scale without jeopardizing patient privacy[36]. Only through such multi-institutional collaboration can robust AI models be trained that make the transition from bench to bedside[36]. Federated learning has been proposed as a solution that allows multiple institutions, individuals, or data providers to collaborate in training AI models without sharing any data with each other[2,37]. This paradigm works well if the data is homogeneously labeled, i.e., if all participating institutions use the same labeling procedure. However, it is the norm rather than the exception that different data providers have similar data but have labeled the data in a seemingly incompatible fashion. Conventional federated learning cannot deal with this situation and new solutions are required. We provide this solution by proposing FFL as a framework for the training on data that is not uniformly labeled. We test this paradigm on a big multi-institutional database of over 680,000 thoracic radiographs from five different hospitals covering the US, Asia and Europe and we find that FFL consistently improves the performance of deep learning models over a wide variety of pathologies.

Our study has limitations. First, we performed all the experiments in a proof-of-concept setup, i.e., within one institutional network, thus the setup is only a simulation of the real situation. However, the setting in which multiple institutions - each with their own network - perform FFL was simulated realistically, by keeping the datasets strictly separate and distributing them to different computing entities. Second, we only tested convolutional neural networks, in particular a ResNet50 architecture. We made that choice to demonstrate our proof-of-concept on one of the most widely used architectures[38–41]. Recently, more general network architectures such as transformers[42–44] have been proposed and may become more important in the future. However, it can be assumed that Transformer architectures may similarly profit from FFL, potentially even stronger than convolutional neural networks since they usually require even bigger data to converge. Third, we only demonstrated FFL for the case of chest radiographs. This is due to the unique availability of public datasets that allow for the study to be performed and to be repeated by other researchers. FFL is not specific to chest X-ray analysis, though. Future works will employ FFL in different domains such as gigapixel imaging in pathology[45,46], and in 3-dimensional volumetric medical imaging such as magnetic resonance imaging and computed tomography.

Our proposed flexible federated learning scheme provides a new way of thinking about collaborative learning. With FFL data does not need to be labeled in an identical fashion at every institution. Rather, machine learning researchers can tap into the vast amount of data that has been labeled heterogeneously and utilize it to train their models on truly big data. This brings secure and privacy-preserving multi-institutional collaboration to the next level and allows the training of models on truly big data.

## Methods

### Ethics statement

The methods were performed in accordance with relevant guidelines and regulations and approved by the ethical committee of the Medical Faculty of RWTH Aachen University. Where necessary, informed consent was obtained from all subjects and/or their legal guardian(s).

### Patient cohorts

VinDr-CXR[18,19] is a cohort containing a total of n=18,000 frontal chest X-ray (CXR) images manually labeled by radiologists. The official training and the benchmark test sets include n=15,000 and n=3,000 images, respectively. The available labels consist of 27 different diseases including *aortic* *enlargement, atelectasis, calcification, cardiomegaly, clavicle fracture, consolidation, edema, emphysema, enlarged pulmonary artery, interstitial lung disease, infiltration, lung opacity, lung cavity, lung cyst, mediastinal shift, nodule/mass, pleural effusion, pleural thickening, pneumothorax, pulmonary fibrosis, rib fracture, other lesion, chronic obstructive pulmonary disease, lung tumor, pneumonia, tuberculosis, other diseases* as well as the *no finding* label.

ChestX-ray14[22] dataset contains a total of n=112,120 frontal x-ray images from 30,805 unique patients[47]. The dataset contains labels for 14 diseases including *atelectasis, cardiomegaly, effusion, infiltration, mass, nodule, pneumonia, pneumothorax, consolidation, edema, emphysema, fibrosis, pleural thickening, hernia* and also for *no finding*. The labels were automatically generated from radiology reports using natural language processing techniques. We adopted the original proposed benchmark test subset including n=25,596 images and utilized the rest of the n=86,524 images as training.

CheXpert[23] dataset v1.0 contains n=224,316 chest radiographs of 65,240 patients. Out of these, 157,676 images are frontal chest radiographs. All the images are automatically labeled based on radiology reports utilizing a natural-language-processing-based labeler. The available labels include *atelectasis, cardiomegaly, consolidation, edema, enlarged cardiomediastinum, fracture, lung lesion, lung opacity, pleural effusion, pleural other, pneumonia, pneumothorax, support devices,* and *no finding*. Unlike ChestX-ray14 and VinDr-CXR datasets which consist of binary labels, CheXpert labels include 4 different classes of "positive", "negative", "uncertain", and "not mentioned in the reports". The "uncertain" label can capture both the uncertainty of a radiologist in the diagnosis as well as ambiguity inherent in the report[23]. We divided the dataset to 80% training and 20% test.

MIMIC Chest X-ray JPG (MIMIC-CXR-JPG) database v2.0.0[24,25] consists of 377,110 CXR images including n=210,652 frontal images for training, 1,691 frontal images for validation, and 2,844 frontal images for test. MIMIC-CXR-JPG provides free-text radiology reports associated with the images. Furthermore, 2 separate sets of labels generated using the labelers from CheXpert[23] and NegBio[48], an open-source rule-based tool for negation and uncertain detection in radiology reports, are provided. We used the labels generated based on the CheXpert labeler in order to be consistent with the CheXpert dataset.

Finally, we employed UKA-CXR[20], a large internal dataset of chest radiographs from RWTH Aachen University Hospital. The dataset consists of n=193,361 frontal CXR images, all manually labeled by the radiologists. The available labels include *pleural effusion*, *pneumonic infiltrates*, *atelectasis*, and *pneumothorax*, each one separately for *right* and *left* parts, and *cardiomegaly*. The labeling system for *cardiomegaly* included 5 classes of "normal", "uncertain", "borderline", "enlarged", and "massively enlarged". For the rest of the labels, 5 classes of "negative", "uncertain", "mild", "moderate", and "severe" were used. Data were split into 75% training and 25% testing data using patient-wise stratification, but otherwise completely random allocation. It is worth noting that, in none of the datasets, there was any overlap between training and test cohorts.

**Data pre-processing**

ChestX-ray14, CheXpert, and MIMIC-CXR-JPG-v2.0 datasets were readily available in PNG standard formats. All the image pixels of the datasets which were only available in digital imaging and communications in medicine (DICOM) format, i.e., VinDr-CXR and UKA-CXR, were extracted and converted into PNG. The DICOM field PhotometricInterpretation was used to determine whether the pixel values were inverted, and if necessary, images were inverted[24]. Only the frontal images were used during the experiments. We followed the same pre-processing scheme for all datasets. All the images were resized to (512 x 512) resolution. Afterwards, a normalization scheme as described before by Johnson et al.[24] was utilized by subtracting the lowest value in the image, dividing by the highest value in the shifted image, truncating values, and converting the result to an unsigned integer, i.e., the range of [0, 255]. Finally, using Python's OpenCV library, histogram equalization was performed by shifting pixel values towards 0 or towards 255[24].

A binary diagnosis paradigm was chosen for all the experiments. ChestX-ray14 and VinDr-CXR datasets included binary labels by design. For the CheXpert dataset (and subsequently for the MIMIC-CXR-JPG-v2.0 dataset), all the 3 classes of "negative", "uncertain", and "not mentioned in the reports" were treated as the negative class and only the original "positive" class was treated as the positive class. For the UKA-CXR dataset, the "negative" and "uncertain" classes ("normal" and "uncertain" for *cardiomegaly*) were treated as negative, while the "mild", "moderate", and "severe" classes ("borderline", "enlarged", and "massively enlarged" for *cardiomegaly*) were treated as positive.

## **Flexible federated learning (FFL) scheme**

The backbone architecture of all networks at all sites was identical by using shared weights of a ResNet50[21]. After each iteration, the locally updated weights were pooled and averaged, and the updated backbone weights were sent back to the sites for the next iteration.

The network head, i.e., the classification layer, was individual to each site and its updates were not aggregated during FFL. This allowed for different classification problems to be backpropagated at each site and made it possible to use data with labels that are unique to each site. For the classification head we employed a fully connected neural network layer as described below. After convergence each site was allowed to perform additional training rounds without central aggregation (i.e., neither of the backbone, nor the classification head) for fine-tuning.

The situation with multiple separate data centers was simulated by isolating each center on a virtual machine within the same network and on the same bare-metal computer. This is slightly different from the real situation in which virtual machines would be set up in different networks but linked through a common virtual private network. However, there is no principal difference to the real setup.

## **Deep learning training procedure**

We performed data augmentation during training by applying medio-lateral flipping with a probability of 0.5 and random rotation in the range of [0, 10] degrees. The ResNet50 architecture was employed as a backbone architecture. We followed the same 50-layer implementation proposed by He et al.[21], where the first layer included a $(7 \times 7)$ convolution producing an output image with 64 channels. The inputs to the network were $(512 \times 512 \times 3)$ images in batches of size 16. Last layer included a linear layer which reduced the $(2048 \times 1)$ output feature vectors to the desired number of diseases to be predicted for each case. The sigmoid function was utilized to convert the output predictions to individual class probabilities. The full network contained a total of 23,512,130 trainable parameters.

All models were optimized using the Adam[49] optimizer. During FFL training of the backbone, a learning rate of $5 \times 10^{-5}$ was chosen. Whereas a learning rate of $9 \times 10^{-5}$ was selected for the training of individual classification heads. As loss function, we chose the binary weighted cross-entropy with inverted class frequencies of the training data as loss weights. It is worth mentioning that even though in our implementation the choice of the loss function type was the same in all networks, as the objectives were not the same, every classification head had an independent loss function.

## Quantitative evaluation

The area under the receiver-operator-curve (AUROC) was used as the primary evaluation metric. Accuracy, sensitivity, and specificity were utilized as further evaluation metrics. We reported the average AUROC over all the labels for each experiment, while the individual AUROC of different labels, as well as accuracy, sensitivity, and specificity are reported in the supplemental material (see **Tables S3-S5**). It should be noted that we followed a multilabel classification paradigm, where multiple diseases could have positive labels given an image. Therefore, we optimized the average performance of the networks over all the diseases, as opposed to optimizing per disease.

## Statistical analysis

Bootstrapping was employed with 1,000 redraws for each measure to determine the statistical spread and calculate p-values for differences[50]. For the calculation of sensitivity and specificity scores, a threshold was chosen according to Youden's criterion[51], i.e., a threshold that maximized (true positive rate - false positive rate).

## Code availability

All source codes for training and evaluation of the deep neural networks, collaborative learning, data augmentation, CXR image analysis, and preprocessing is publicly available at https://github.com/tayebiarasteh/chestx. All code for the experiments was developed in Python 3.8 using the PyTorch 1.4 framework. The collaborative learning process was developed using PySyft 0.2.9[52].

## Data availability

ChestX-ray14 data is publicly available under https://www.v7labs.com/open-datasets/chestx-ray14. VinDr-CXR and MIMIC-CXR-JPG data are restricted-access resources, which can be accessed from PhysioNet by agreeing to its data protection requirements under https://physionet.org/content/vindr-cxr/1.0.0/ and https://physionet.org/content/mimic-cxr-jpg/2.0.0/, respectively. CheXpert data can be requested from Stanford University at https://stanfordmlgroup.github.io/competitions/chexpert/. The UKA-CXR data is not publicly accessible as it is internal data of patients of the University Hospital RWTH Aachen. A reasonable request from the corresponding author is required for accessing the data.

# Additional information

## Acknowledgements

The authors would like to thank the data providers of VinDr-CXR, ChestX-ray14, CheXpert, and MIMIC-CXR-JPG for providing them access to their public datasets. We additionally acknowledge the support by NVIDIA who provided our group with two RTX6000 GPUs.

## Funding sources

This work was (partially) funded / supported by the RACOON network under BMBF grant number 01KX2021. JNK is supported by the German Federal Ministry of Health (DEEP LIVER, ZMVI1-2520DAT111) and the Max-Eder-Programme of the German Cancer Aid (grant #70113864), the German Federal Ministry of Education and Research (PEARL, 01KD2104C), and the German Academic Exchange Service (SECAI, 57616814).

## Author contributions

The formal analysis was conducted by STA and DT and the original draft was written by STA and reviewed and corrected by DT and SN. The software was mainly developed by STA. FK, GMF, JNK and DT provided technical expertise, PI, MS, JNK, CK, SN and DT provided clinical expertise. The experiments were performed by STA. All authors read the manuscript and contributed to the interpretation of the results and agreed to the submission of this paper.

## Competing interests

JNK reports consulting services for Owkin, France, Panakeia, UK and DoMore Diagnostics, Norway and has received honoraria for lectures by MSD, Eisai and Fresenius. The other authors do not have any competing interests to disclose.

## Hardware

The hardware used in our experiments were Intel CPUs with 18 cores and 32 GB RAM and Nvidia RTX 6000 GPUs with 24 GB memory.

## Correspondence

Daniel Truhn, University Hospital RWTH Aachen, Pauwelsstr. 30, 52074 Aachen, Germany, Email: dtruhn@ukaachen.de.

# Supplemental Information

## **Further Information on the Utilized Datasets**

**Tables S1** and **S2** show the demographics of the utilized datasets in this study as well as the different imaging findings available by each of them.

**Table S1**: Statistics of the available chest radiograph datasets utilized in this study. Automatic labeling system means the labels were generated using automatic natural-language-processing-based systems from radiology reports whereas manual means the images were manually labeled by radiologists.

|  | VinDr-CXR[18,19] | ChestX-ray14[22] | CheXpert[23] | MIMIC-CXR[24,25] | UKA-CXR[20] |
|---|---|---|---|---|---|
| Total training frontal images | 15,000 | 86,524 | 128,356 | 210,652 | 122,294 |
| Total test frontal images | 3,000 | 25,596 | 29,320 | 2,844 | 39,824 |
| Total frontal images | 18,000 | 112,120 | 157,676 | 215,187 | 193,361 |
| Number of labels available | 28 | 15 | 14 | 14 | 9 |
| Labeling system for training set | manual | automatic | automatic | automatic | manual |
| Labeling system for test set | manual | automatic | automatic | automatic | manual |

**Table S2**: Details of available labels corresponding to each chest radiograph dataset utilized in this study. "No finding" indicates the absence of all other diseases.

| Label name | VinDr-CXR[18,19] | ChestX-ray14[22] | CheXpert[23] | MIMIC-CXR[24,25] | UKA-CXR[20] |
|---|---|---|---|---|---|
| *Aortic enlargement* | yes | No | no | no | no |
| *Atelectasis* | yes | yes | yes | yes | yes |
| *Calcification* | yes | no | no | no | no |
| *Cardiomegaly* | yes | yes | yes | yes | yes |
| *Clavicle fracture* | yes | no | yes | yes | no |
| *Consolidation* | yes | yes | yes | yes | no |
| *Edema* | yes | yes | yes | yes | no |
| *Emphysema* | yes | yes | no | no | no |
| Enlarged cardiomediastinum | no | no | yes | yes | no |
| *Enlarged PA* | yes | no | no | no | no |
| *Hernia* | no | yes | no | no | no |
| *Interstitial lung disease* | yes | no | no | no | no |
| *Infiltration* | yes | yes | no | no | yes |
| *Lung cavity* | yes | no | no | no | no |
| *Lung lesion* | yes | no | yes | yes | no |
| *Lung opacity* | yes | no | yes | yes | no |
| *Lung tumor* | yes | no | no | no | no |
| *Mediastinal shift* | yes | no | no | no | no |
| *No finding* | yes | yes | yes | yes | no |
| *Nodule/Mass* | yes | yes | no | no | no |
| *Pleural effusion* | yes | yes | yes | yes | yes |
| *Pleural thickening* | yes | yes | yes | yes | no |
| *Pneumonia* | yes | yes | yes | yes | yes |
| *Pneumothorax* | yes | yes | yes | yes | yes |
| *Pulmonary fibrosis* | yes | yes | no | no | no |
| *Rib fracture* | yes | no | yes | yes | no |
| *Tuberculosis* | yes | no | no | no | no |

# Detailed Evaluation Results of All Experiments

Detailed evaluation results of all experiments in our study for all the individual diseases including further evaluation metrics are provided in **Tables S3–S5**.

**Table S3**: Details of the results of the comparison between local and FFL-based training of VinDr-CXR dataset with non-overlapping labels for different training set sizes, tested on the VinDr-CXR test set. The results show the individual AUROC values for each label, and average accuracy, sensitivity, and specificity values over all labels. The FFL was performed in combination with UKA-CXR dataset of n=122,294 images with 2 other labels including *pleural effusion right* and *pneumonic infiltrates left*.

|  | Local 2K | Local 5K | Local 15K | FFL 2K | FFL 5K | FFL 15K |
|---|---|---|---|---|---|---|
| *Cardiomegaly* | 0.90 ± 0.01 | 0.89 ± 0.01 | 0.92 ± 0.01 | 0.88 ± 0.01 | 0.90 ± 0.01 | 0.94 ± 0.01 |
| *Pleural effusion* | 0.82 ± 0.02 | 0.90 ± 0.02 | 0.96 ± 0.01 | 0.91 ± 0.02 | 0.94 ± 0.01 | 0.96 ± 0.01 |
| Average accuracy | 0.78 ± 0.08 | 0.85 ± 0.03 | 0.89 ± 0.04 | 0.84 ± 0.07 | 0.88 ± 0.05 | 0.90 ± 0.03 |
| Average sensitivity | 0.80 ± 0.10 | 0.80 ± 0.04 | 0.87 ± 0.06 | 0.80 ± 0.06 | 0.81 ± 0.05 | 0.89 ± 0.03 |
| Average specificity | 0.77 ± 0.09 | 0.86 ± 0.04 | 0.89 ± 0.04 | 0.84 ± 0.08 | 0.88 ± 0.05 | 0.90 ± 0.03 |

**Table S4**: Results of the comparison between local and FFL-based training of VinDr-CXR dataset with overlapping labels for different training set sizes, tested on the VinDr-CXR test set. Results show the individual AUROC values for each label, and average accuracy, sensitivity, and specificity values over all labels. The FFL was performed in combination with UKA-CXR dataset of n=122,294 images with 7 other labels including *cardiomegaly*, *pleural effusion right*, *pleural effusion left*, *pneumonic infiltrates right*, *pneumonic infiltrates left*, *atelectasis right*, and *atelectasis left*.

|  | Local 2K | Local 5K | Local 15K | FFL 2K | FFL 5K | FFL 15K |
|---|---|---|---|---|---|---|
| *No finding* | 0.81 ± 0.01 | 0.79 ± 0.01 | 0.83 ± 0.01 | 0.80 ± 0.01 | 0.83 ± 0.01 | 0.83 ± 0.01 |
| *Aortic enlargement* | 0.82 ± 0.01 | 0.82 ± 0.01 | 0.83 ± 0.01 | 0.81 ± 0.01 | 0.83 ± 0.01 | 0.85 ± 0.01 |
| *Pleural thickening* | 0.71 ± 0.02 | 0.75 ± 0.02 | 0.81 ± 0.02 | 0.77 ± 0.02 | 0.78 ± 0.02 | 0.77 ± 0.02 |
| *Cardiomegaly* | 0.86 ± 0.01 | 0.86 ± 0.01 | 0.90 ± 0.01 | 0.86 ± 0.01 | 0.89 ± 0.01 | 0.90 ± 0.01 |
| *Pleural effusion* | 0.84 ± 0.02 | 0.89 ± 0.02 | 0.95 ± 0.01 | 0.80 ± 0.02 | 0.85 ± 0.02 | 0.90 ± 0.02 |
| *Pneumothorax* | 0.74 ± 0.06 | 0.70 ± 0.07 | 0.85 ± 0.06 | 0.71 ± 0.07 | 0.79 ± 0.06 | 0.86 ± 0.04 |
| *Atelectasis* | 0.63 ± 0.03 | 0.71 ± 0.03 | 0.65 ± 0.03 | 0.70 ± 0.03 | 0.77 ± 0.03 | 0.79 ± 0.03 |
| Average accuracy | 0.74 ± 0.08 | 0.74 ± 0.08 | 0.76 ± 0.12 | 0.74 ± 0.07 | 0.77 ± 0.06 | 0.77 ± 0.08 |
| Average sensitivity | 0.71 ± 0.12 | 0.71 ± 0.11 | 0.78 ± 0.09 | 0.71 ± 0.15 | 0.75 ± 0.09 | 0.77 ± 0.09 |
| Average specificity | 0.74 ± 0.08 | 0.74 ± 0.09 | 0.75 ± 0.13 | 0.74 ± 0.08 | 0.76 ± 0.07 | 0.77 ± 0.09 |

**Table S5**: Details of the results of the comparison between local and FFL-based training for 5 different datasets. The results show the individual AUROC values for each label, and average accuracy, sensitivity, and specificity values over all labels, tested on the test benchmarks of the corresponding datasets. The FFL process for each dataset was performed in combination with the other 4 datasets including 5 different labels for each dataset. VinDr-CXR, ChestX-ray14, CheXpert, MIMIC-CXR, and UKA-CXR cohorts contained n=15,000, n=86,524, n=128,356, n=210,652, and n=122,294 training images, respectively.

| Dataset name | | Individual AUROC (in order) | Average accuracy | Average sensitivity | Average specificity |
|---|---|---|---|---|---|
| **VinDr-CXR** *(no finding, aortic enlargement, pleural thickening, cardiomegaly, pleural effusion)* | Local | 0.841 ± 0.008, 0.859 ± 0.011, 0.810 ± 0.017, 0.893 ± 0.011, 0.933 ± 0.013 | 0.777 ± 0.065 | 0.822 ± 0.049 | 0.763 ± 0.074 |
| | FFL | 0.849 ± 0.008, 0.869 ± 0.011, 0.832 ± 0.016, 0.909 ± 0.009, 0.966 ± 0.007 | 0.816 ± 0.087 | 0.823 ± 0.050 | 0.801 ± 0.102 |
| **ChestX-ray14** *(cardiomegaly, effusion, pneumonia, consolidation, no finding)* | Local | 0.865 ± 0.005, 0.791 ± 0.003, 0.653 ± 0.011, 0.709 ± 0.006, 0.699 ± 0.003 | 0.655 ± 0.095 | 0.718 ± 0.093 | 0.661 ± 0.106 |
| | FFL | 0.874 ± 0.05, 0.793 ± 0.003, 0.645 ± 0.011, 0.705 ± 0.006, 0.705 ± 0.003 | 0.658 ± 0.079 | 0.710 ± 0.105 | 0.667 ± 0.095 |
| **CheXpert** *(cardiomegaly, lung opacity, lung lesion, pneumonia, edema)* | Local | 0.877 ± 0.003, 0.723 ± 0.003, 0.771 ± 0.007, 0.743 ± 0.009, 0.866 ± 0.002 | 0.722 ± 0.059 | 0.759 ± 0.064 | 0.697 ± 0.087 |
| | FFL | 0.876 ± 0.003, 0.726 ± 0.003, 0.769 ± 0.007, 0.751 ± 0.009, 0.863 ± 0.002 | 0.718 ± 0.063 | 0.762 ± 0.062 | 0.695 ± 0.086 |
| **MIMIC-CXR** *(enlarged cardiomediastinum, consolidation, pleural effusion, pneumothorax, atelectasis)* | Local | 0.701 ± 0.025, 0.704 ± 0.019, 0.876 ± 0.007, 0.831 ± 0.020, 0.749 ± 0.010 | 0.704 ± 0.100 | 0.723 ± 0.081 | 0.702 ± 0.110 |
| | FFL | 0.718 ± 0.026, 0.734 ± 0.018, 0.876 ± 0.007, 0.848 ± 0.021, 0.754 ± 0.010 | 0.716 ± 0.095 | 0.744 ± 0.106 | 0.710 ± 0.106 |
| **UKA-CXR** *(pleural effusion left, pleural effusion right, cardiomegaly, pneumonic infiltrates left, pneumonic infiltrates right)* | Local | 0.920 ± 0.003, 0.940 ± 0.002, 0.855 ± 0.002, 0.935 ± 0.002, 0.930 ± 0.002 | 0.824 ± 0.031 | 0.860 ± 0.046 | 0.818 ± 0.036 |
| | FFL | 0.925 ± 0.003, 0.942 ± 0.002, 0.857 ± 0.002, 0.937 ± 0.002, 0.932 ± 0.002 | 0.832 ± 0.035 | 0.858 ± 0.038 | 0.825 ± 0.044 |

**Exemplary Radiographs from the Utilized Datasets**

**Fig. S1** illustrative radiographs from the internal UKA-CXR[20] dataset, utilized in this study. All other data, including the images and cohort statistics (VinDr-CXR[18,19], ChestX-ray14[22], CheXpert[23], and MIMIC-CXR[24,25]) are publicly available.

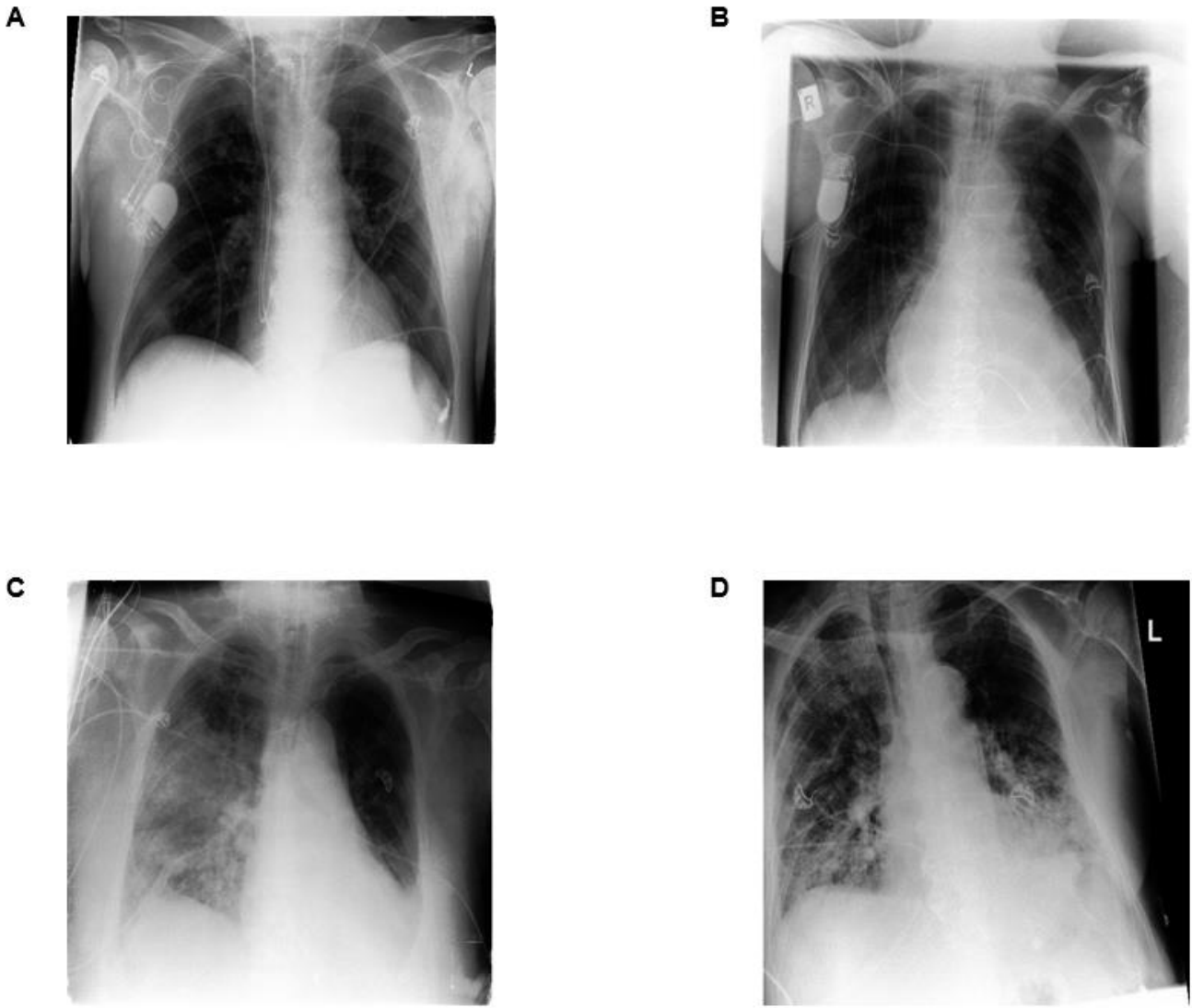

**Fig. S1**: Exemplary radiographs utilized in this study. **(A)** 70-year-old male healthy subject (radiologists reported "no finding", i.e., no pathological changes. Cardiac pacemaker can be seen on the right patient side). **(B)** 86-year-old male patient diagnosed with *cardiomegaly*. **(C)** 71-year-old male patient diagnosed with *cardiomegaly*, *atelectasis right*, and *pneumonic infiltrates right*. **(D)** 86-year-old male patient diagnosed with *atelectasis right*, *atelectasis left, pneumonic infiltrates right,* and *pneumonic infiltrates left.*

## Exemplary Convergence and ROC Curves

**Fig. S2** and **Fig. S3** show exemplary training loss and receiver-operator-characteristic (ROC) curves. FFL-based training is performed with partially overlapping labels on UKA-CXR[20] (n=122,294, labels: *cardiomegaly*, *pleural effusion right*, *pleural effusion left*, *pneumonic infiltrates right*, *pneumonic infiltrates left*, *atelectasis right*, and *atelectasis left*) and on VinDr-CXR[18,19] (on 2K images, labels: over *no finding*, *aortic enlargement*, *pleural thickening*, *cardiomegaly*, *pleural effusion*, *pneumothorax*, and *atelectasis*). Performance tested on an independent VinDr-CXR test set.

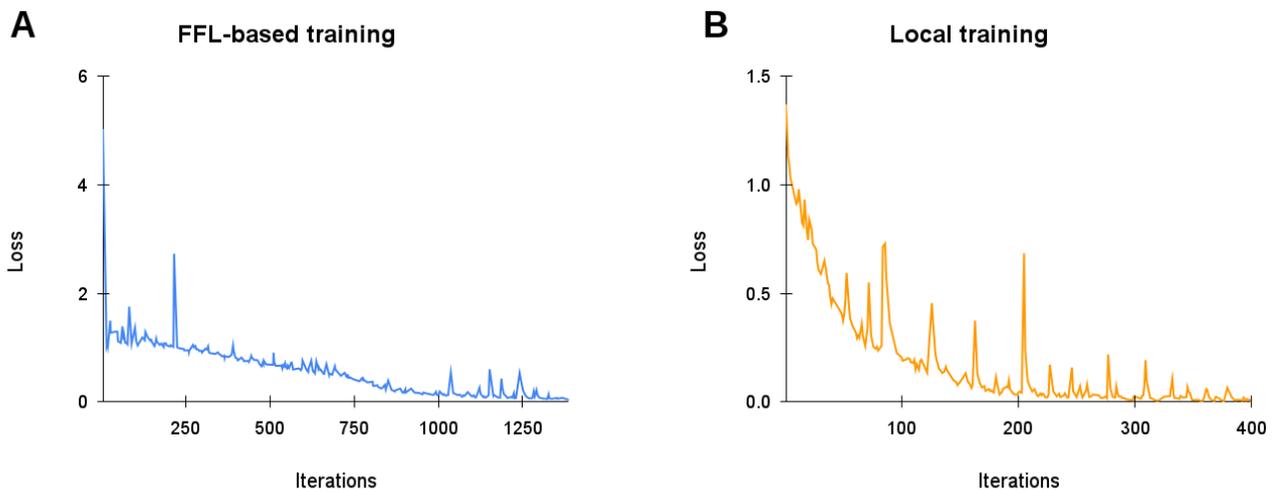

**Fig. S2**: Training loss curves for FFL-based training and local training of classification models. FFL-based training is performed with partially overlapping labels on UKA-CXR[20] (n=122,294, labels: *cardiomegaly*, *pleural effusion right*, *pleural effusion left*, *pneumonic infiltrates right*, *pneumonic infiltrates left*, *atelectasis right*, and *atelectasis left*) and on VinDr-CXR[18,19] (on 2K images, labels: over *no finding*, *aortic enlargement*, *pleural thickening*, *cardiomegaly*, *pleural effusion*, *pneumothorax*, and *atelectasis*). **(A)** FFL-based training, **(B)** Local training.

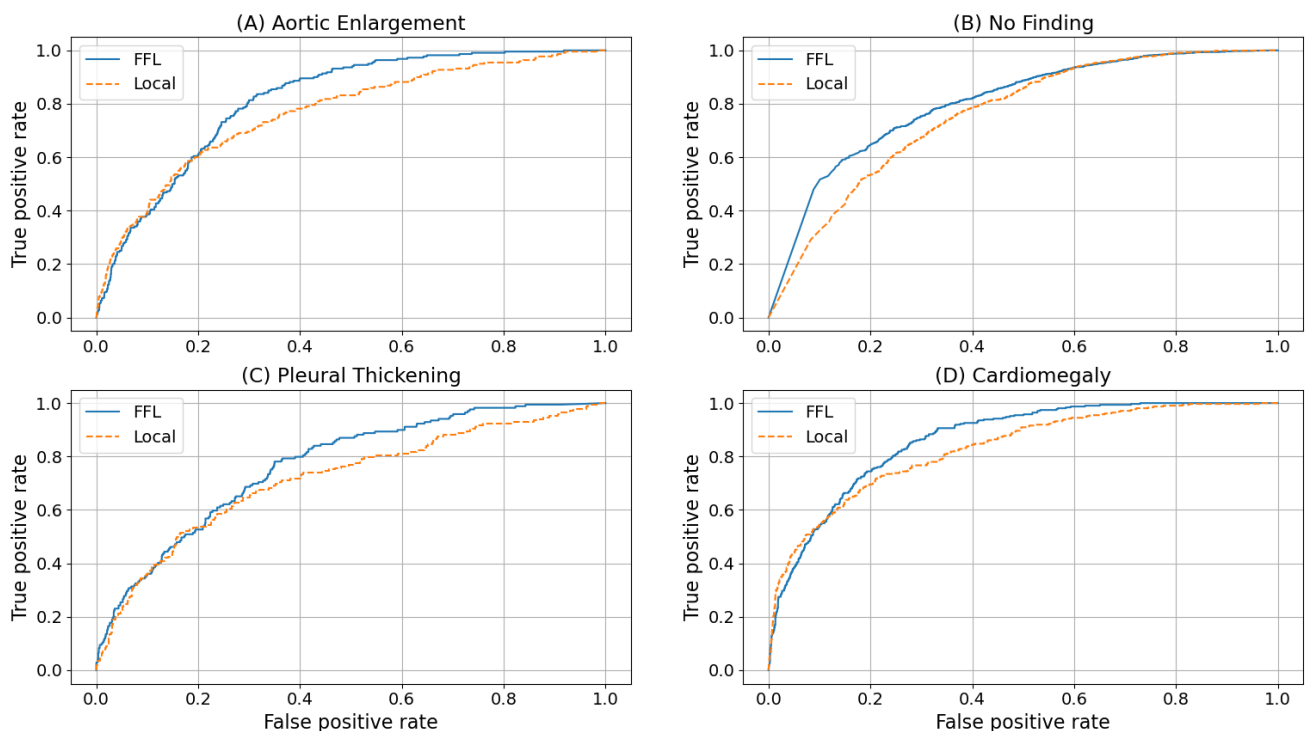

**Fig. S3**: Comparison between FFL-based training and local training of classification models in terms of receiver-operator-characteristic (ROC) curves. FFL-based training is performed with partially overlapping labels on UKA-CXR[20] (n=122,294, labels: *cardiomegaly*, *pleural effusion right*, *pleural effusion left*, *pneumonic infiltrates right*, *pneumonic infiltrates left*, *atelectasis right*, and *atelectasis left*) and on VinDr-CXR[18,19] (on 2K images, labels: over *no finding*, *aortic enlargement*, *pleural thickening*, *cardiomegaly*, *pleural effusion*, *pneumothorax*, and *atelectasis*). Performance tested on an independent VinDr-CXR test set. ROC curves are illustrated for individual labels including **(A)** aortic enlargement, **(B)** *no finding*, **(C)** pleural thickening, and **(D)** *cardiomegaly*.